\documentclass[journal]{IEEEtran}
\ifCLASSINFOpdf
\else
\fi
\usepackage{graphicx}
\usepackage{float}
\usepackage{amssymb}
\usepackage{amsmath}
\usepackage{cite}
\usepackage[switch]{lineno}
\usepackage{multirow}
\usepackage{stfloats}
\usepackage{algorithm, algorithmic}

\begin{document}
\title{Incomplete Multimodal Learning for Remote Sensing Data Fusion}
%
\author{Yuxing~Chen,~\IEEEmembership{Graduate Student,~IEEE}
		Maofan~Zhao,~\IEEEmembership{Graduate Student,~IEEE}
	    Lorenzo~Bruzzone,~\IEEEmembership{Fellow,~IEEE}
        
\thanks{Y. Chen and L. Bruzzone are with the Department of Information Engineering and Computer Science, University of Trento, 38122 Trento, Italy (e-mail:yuxing.chen@unitn.it;lorenzo.bruzzone@unitn.it).}
\thanks{Corresponding author: L. Bruzzone}}

\maketitle

\begin{abstract}
The mechanism of connecting multimodal signals through self-attention operation is a key factor in the success of multimodal Transformer networks in remote sensing data fusion tasks.
However, traditional approaches assume access to all modalities during both training and inference, which can lead to severe degradation when dealing with modal-incomplete inputs in downstream applications.
To address this limitation, our proposed approach introduces a novel model for incomplete multimodal learning in the context of remote sensing data fusion.
This approach can be used in both supervised and self-supervised pretraining paradigms and leverages the additional learned fusion tokens in combination with Bi-LSTM attention and masked self-attention mechanisms to collect multimodal signals.
The proposed approach employs reconstruction and contrastive loss to facilitate fusion in pre-training while allowing for random modality combinations as inputs in network training.
Our approach delivers state-of-the-art performance on two multimodal datasets for tasks such as building instance / semantic segmentation and land-cover mapping tasks when dealing with incomplete inputs during inference.
\end{abstract}


\section{Introduction}
\IEEEPARstart{R}{emote} sensing becomes more and more important in various Earth Observation (EO) tasks due to its superior ability in observing our planet.
With the increasing availability of multimodal RS data, researchers now can develop more diverse downstream applications.
Despite the abundance of multimodal remote sensing data, each modality captures only certain specific properties and, therefore, cannot thoroughly describe the observed scenes, which poses a great constraint on subsequent applications.
Multimodal RS data fusion proves to be a feasible solution to overcome the inadequacy posed by unimodal data \cite{ghamisi2019multisource}.
Different modal data provide complementary information due to their varying perspectives, allowing the researchers to conduct extensive research on combining useful information from different modalities to better achieve specific application goals.
For instance, synthetic aperture radar (SAR) provides physical structure information, while LiDAR collects depth information \cite{paris2014three}.
Meanwhile, multispectral (MS) and hyperspectral (HS) data measure radiation reflectance across different wavelengths of the electromagnetic spectrum.
By integrating the complementary information from multimodal data, researchers can make a more robust and reliable model in various tasks, such as change detection \cite{chen2021self} and land-cover mapping \cite{chen2022approach}.
To integrate the complementary information provided by these modalities, such as MS, HS and SAR, as well as the remote sensing products (e.g., Land Cover Land Use Maps), traditional methods have been intensively studied by designing handcrafted features based on domain-specific knowledge and exploiting rough fusion strategies, which inevitably impairs the fusion performance, especially for heterogeneous data.
For example, Deus \cite{deus2016integration} tried to design some features manually (i.e. NDVI, RFDI, texture, etc.) based on ALOS PALSAR and Landsat TM images to improve the performance of support vector machines on land cover classification and forest mapping. 

Thanks to the growth of artificial intelligence, deep learning shows great potential in modelling a complex relationship between different modality data and is widely used in remote sensing data fusion tasks.
However, most of them are based on the supervised learning paradigm.
Supervised approaches are task-specific and cannot be generalized to other tasks.
However, training on a large amount of multimodal data is cost expensive and collecting adequate multimodal data for each task is challenging for end-users.
Thus, the research community seeks a few finetuning steps on a pre-trained model, which can help downstream tasks. 
Pretraining without supervision has gained a lot of attention as it is more general and can lead to improvements.
Self-supervised learning for SAR-optical feature fusion proposed by Chen et al. \cite{chen2021self} is an example of such an approach.
Nevertheless, both supervised and unsupervised methods assume that all modalities are available during training and inference, which can be a limiting factor in practical applications, as data collection processes may miss modalities. 
In such cases, existing multimodal data fusion methods may fail to deal with incomplete modalities, leading to severe degradation in downstream tasks. 
As a solution, a robust multimodal method is needed for flexible and practical RS applications, with or without missing modalities.
The algorithm used in this situation is called incomplete multimodal learning, which aims at learning methods that are robust with any subset of available modalities at inference.
A simple strategy for incomplete multimodal learning is to synthesize the missing modalities using generative models. 
For instance, Generative Adversarial Networks (GANs) can effectively overcome the problems arising from missing or incomplete modalities in building footprint segmentation, as proposed by Bischke et al \cite{bischke2018overcoming}. 
Another set of methods explores knowledge distillation from complete to incomplete modalities. 
In this approach, Kampffmeyer et al. \cite{kampffmeyer2018urban} proposed to use an additional network, the hallucination network, for when data modalities are missing during the testing of Urban Land Cover Classification tasks. 
The network takes a modality as input that is assumed to be available during both training and testing, trying to learn a mapping function from this modality to the missing modality.

Although promising results are obtained, such methods have to train and deploy a specific model for each subset of missing modalities, which is complicated and burdensome in downstream tasks.
Meanwhile, all these methods require complete modalities during the training process.
Recent incomplete multimodal learning methods focus on learning a unified model, instead of a bunch of distilled networks, for downstream tasks.
In this context, the modality-invariant fusion embedding across different modalities may contribute to more robust performance, especially when one or more modalities are missing.
Transformer is widely used in this task for its flexibility and multimodality modelling abilities.
Current works exploited Transformers for audio and video data fusion using contrastive learning \cite{nagrani2021attention, recasens2023zorro}.
However, the dedicated Transformer for incomplete multimodal learning of remote sensing tasks has not been carefully tapped yet and the existing multimodal RS data fusion methods cannot allow missing data in the training process.
This paper proposes to exploit Transformer to build a unified model for incomplete multimodal learning of remote sensing tasks, which can be used in both the supervised paradigm and self-supervised pre-training paradigm.
This is achieved by using additional learned fusion tokens for multimodal signal collection in the network.
However, only using the additional learned fusion token cannot query enough information from other modality tokens.
In this context, we use the Bi-LSTM attention block to further distil different modality information to fusion tokens.
Using this algorithm, the proposed approach can leverage MultiMAE and contrastive loss to build fusion across the different modalities in pre-training.
And also make it possible to use a random modality combination training strategy in downstream task finetuning.
All these make the learning and inference of the modal can be in an incomplete modality input.

This paper presents three contributions: (1) we propose to use Bi-LSTM and masked self-attention in multimodal Transformer to help build additional fusion tokens across different modalities, which enable both contrastive and generative self-supervised pre-training for incomplete multimodal inputs;
(2) based on the proposed strategies, we use the random modality combination training strategy in downstream tasks, which ensures task performance with incomplete inputs on inference.
(3) we benchmark our approach on two datasets: DFC2023 track2 and created quadruplet dataset, which shows the proposed approach can be pre-trained on a large-scale remote sensing multimodal dataset in a self-supervised manner.
The proposed approach achieves state-of-the-art performance when compared with the vallina multimodal Transformer on RS.

The rest of this paper is organized as follows.  
Section II presents the related works of multimodal RS data fusion, Masked Autoencoder and multimodal Transformer.   
Section III introduces the proposed approach by describing the network architecture, Bi-LSTM attention, masked self-attention, reconstruction pretraining and contrastive pretraining as well as the random modality combination training strategy.
The descriptions of the dataset, network setup, experimental settings and downstream tasks are given in Section IV.
Experimental results obtained on building instance/semantic segmentation and LULC mapping tasks are also illustrated in Section IV. 
Finally, Section VI concludes the paper.

\section{Related Works}
\subsection{Multimodal RS Data Fusion}
In recent years, deep learning methods have been widely adopted in multimodal RS data fusion, including LiDAR-optical\cite{paris2014three,seydi2022bdd,zhang2018feature,diab2022deep}, SAR-optical\cite{schmitt2018sen1, adrian2021sentinel, adrian2021sentinel, ienco2019combining, kussul2017deep}, image-map fusion \cite{kemper2020sensor,xu2023road}.
In the case of LiDAR-optical data fusion, Paisitkriangkrai et al. \cite{paisitkriangkrai2015effective} proposed fusing optical and Lidar data through concatenating deep and expert features as inputs to random forests.
Several advanced techniques have subsequently been developed, with the aim of enhancing feature extraction ability.
Audebert et al. \cite{audebert2018beyond} suggest the use of deep fully convolutional networks to investigate the early and late fusion of LiDAR and multispectral data.
Similarly, Chen et al. \cite{chen2017deep} employ a two-branch network to separately extract spectral-spatial-elevation features, followed by utilizing a fully connected layer to integrate these heterogeneous features for final classification.
Other novel fusion strategies are also introduced, such as cross-attention module \cite{mohla2020fusatnet}, a reconstruction-based network \cite{hong2020deep}, and a graph fusion network \cite{du2021multisource}.
Additionally, recent studies by Roy et al. \cite{roy2022multimodal} propose a multimodal Transformer network to fuse Lidar and hyperspectral images for classification.
Similar to Lidar-optical fusion, many researchers also developed the DSM-optical fusion methods, where the DSM was acquired from stereo-optical images.
Meanwhile, SAR-optical data fusion also gets a lot of attention and widely adopts deep learning methods.
For example, Kussul \textit{et al.}\cite{kussul2017deep} first explores the deep CNNs in SAR-optical fusion for LULC classification and demonstrates their superiority with respect to traditional MLP classifiers.
Recent studies by Dino \textit{et al.} \cite{ienco2019combining} propose a deep learning architecture, namely TWINNS, to fuse Sentinel-1 and Sentinel-2 time series data in land-cover mapping.
Similarly, Adrian \textit{et al.} \cite{adrian2021sentinel} use the 3-dimensional deep learning network to fuse multi-temporal Sentinel-1 and Sentinel-2 data for mapping ten different crop types, as well as water, soil and urban area.
Map data, such as topography, land use, road and census data, may be combined with remotely sensed data to improve the accuracy of image classification, object recognition, and change detection.  
For example, Sun et al. \cite{sun2010data} provides a method of data fusion of GIS and RS using a neural network with unchanging data memory structure based on users’ aim. 
Xu et al \cite{xu2023road} propose to conduct road extraction based on satellite images and partial road maps using a two-branch partial to-complete network.

\subsection{Masked Autoencoder}
The MAE (masked autoencoder) \cite{he2022masked} is a novel self-supervised learning algorithm that demonstrates state-of-the-art performance on various vision benchmarks.
Instead of relying on a contrastive objective, the MAE utilizes a pretext task that involves reconstructing masked patches of the input.

The MAE network follows an asymmetric encoding and decoding scheme.
Suppose the input image is a tensor of dimensions $I \in R^{C \times H \times W}$, where $H, W$ are the height and width of the image, respectively, and $C$ is the number of channels.
The image is initially divided into non-overlapping patches $S \in R^{L\times P^2C}$, where $P$ is the height and width of the patch, and $L=(H/P) \times (W/P)$ is the number of patches.
These patches are then transformed into a sequence of embedded patch tokens $S' \in R^{L \times D}$, using a patch embedding function $f_p: R^{P^2C} \rightarrow R^D$.
A fraction $p_m$ of the sequence tokens is randomly masked, and the remaining visible tokens are fed into an encoder, which is a Vision Transformer (ViT).
Due to the lack of positional information, additional positional embeddings are then added to patch embeddings to capture the spatial location of the patch in the image.
The decoder is composed of multiple transformer blocks that are trained for all tokens, where the masked tokens are replaced as the initialized learnable tokens.
The decoder produces a reconstructed image, which is compared to the original image using mean-squared error (MSE) loss, computed only on masked patches.
Positional encoding allows the transformer to encode positional information.
In MAE the positional encoding is:
\begin{equation}
	\operatorname{Encode}(k, 2 i)=\sin \frac{k}{\Omega^{\frac{2 i}{d}}}, \operatorname{Encode}(k, 2 i+1)=\cos \frac{k}{\Omega^{\frac{2 i}{d}}}
\end{equation}
Here, $k$ is the position, $i$ is the index of feature dimension in the encoding, $d$ is the number of possible positions, and $\Omega$ is a large constant. 
In MAE, the position is defined as the index of the patch along the $x$ or $y$ axis.
Therefore, $k$ ranges from $0$ to $H/P$ or $W/P$.
This encoding provides two unique dimensions, one for $x$ and one for $y$ coordinates, which are concatenated for the final encoding representation.

The Multimodal Masked Autoencoder (MultiMAE) \cite{bachmann2022multimae} utilizes a standard single-modal ViT and the encoder.
The encoder is equipped with 2-D sine-cosine positional embeddings following the linear projection.
MultiMAE does not make use of modality-specific embeddings.
Because the bias term in each linear projection is sufficient.
MultiMAE employs a separate decoder for each task that is responsible for reconstructing the masked-out tokens from the visible tokens.
The input to each decoder is a full set of visible tokens from all different modalities, including the learnable modality embeddings with 2-D sine-cosine positional embeddings.
The input is then followed by MLPs and Transformer blocks.
Only the masked tokens are considered in the loss calculation.
The mask sampling strategy employed in MultiMAE plays a crucial role in achieving predictive coding across different modalities.
This sampling strategy ensures that most modalities are represented to similar degrees.
MultiMAE adopts a symmetric Dirichlet distribution to select the proportion of tokens per modality $\lambda$ ($\lambda_i ~ Dir(\alpha)$), where $\sum \lambda_i = 1, \lambda > 0$.
The concentration parameter $\alpha > 0$ controls the sampling.
For simplicity and better representation parameter $\alpha =1$ in MultiMAE.

\subsection{Multimodal Transformer} 
The self-attention blocks of Transformers build a natural bridge among multimodal signals in a unified architecture.
Different from the Convolutional Neural Networks (CNNs) using different network for different modalities, the Transformer only use the same main architecture for all modalities with a modal-specific projector.
Transformers integrate input tokens from all modalities into a single representation, while CNNs fuse features of each modality through concatenation or tensor fusion.
However, such explicit integration necessitates the presence of all modalities during training, which undermines the pipeline in case of a missing modality.
In contrast, Transformers use self-attention to embed a holistic multimodal representation and handle the absence of modalities by applying a mask on the attention matrix. 
Thus, multimodal Transformers are more adaptable to deal with modal-incomplete inputs.
In addition, an easy-to-train model is vital for multimodal learning.
The training load of a conventional multimodal backbone grows as the number of modalities increases since the backbone usually consists of modality-specific sub-models that need to be trained independently for each modality.
Instead, Transformers process all modalities altogether in a single model, significantly reducing the training load.

However, Transformer models exhibit significant deterioration in performance with model-incomplete inputs, especially in the context of multimodal inference where Transformer models tend to overfit to dominate modalities.
To overcome this challenge, MBT \cite{nagrani2021attention} build a multimodal architecture for video and audio, where it uses an additional fusion token to force information among different modalities to pass through by using cross-attention.
However, the representation of each modality can also access each other in MBT, which means they are not independent.
Furthermore, Zorro \cite{recasens2023zorro} employs a modality-aware masking mechanism in all attention operations to isolate the allocation of latent representations of individual modalities, which leads to a resultant representation that is partially unimodal (i.e., part of the representation attends to a single modality) and partially multimodal (i.e., part of the representation attends to all modalities), thereby allowing for the use of contrastive learning.

\begin{figure*}[pt]
	\centering
	\includegraphics[width=7.0 in]{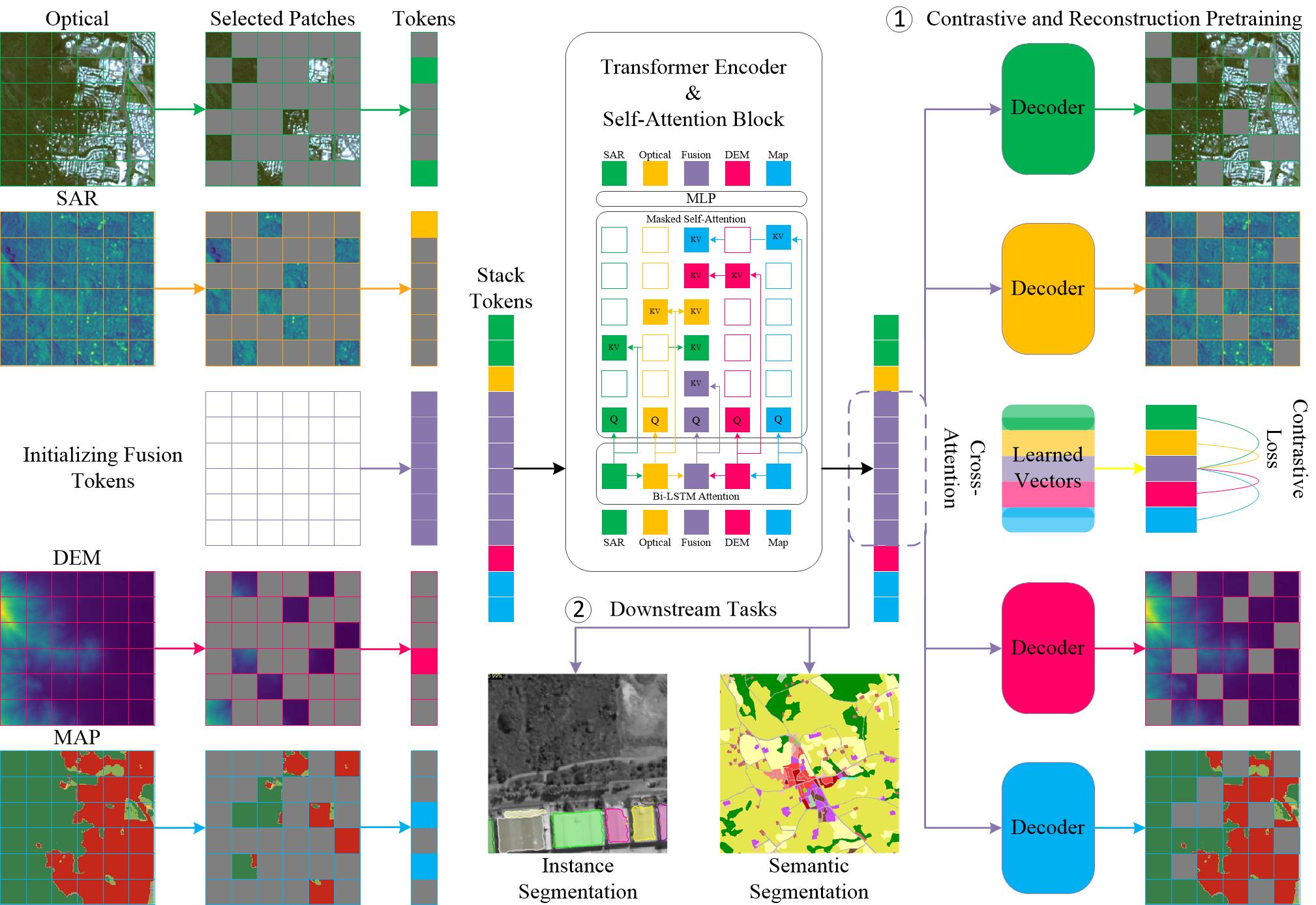}
	\caption{Overview of the proposed framework. The input to our model is optical images, SAR images, DEM and Maps. Each of those inputs is patched using 2D convolution and projected to feature vectors. All inputs are concatenated with a set of learnable fusion tokens and added to the position embedding. Next, we process these inputs through the Transformer Encoder, where the Bi-LSTM Attention and the masked Self-Attention strategy are applied. (1) In pretraining, task-specific decoders reconstruct the masked patches by using the output fusion tokens. Meanwhile, the global vectors of each modality and fusion tokens are output using cross-attention, which allows using contrastive loss between fusion tokens and each modality. (2) In the supervised training, the proposed framework can be trained on downstream tasks by using a random modality combination strategy.}
	\label{fig6-1}
\end{figure*}

\section{Methodology}
In this Section, we describe the incomplete multi-modal fusion architecture with additional learned fusion tokens, Bi-LSTM and masked Self-Attention in an optical-SAR-DEM-MAP data fusion example.
Then, we introduce the details of the pretraining using MultiMAE and contrastive loss, and the details of training using random modality combination on downstream tasks (as shown in Fig. \ref{fig6-1}.

\subsection{Network Architecture}
The main architecture of the proposed approach is a ViT with modality-specific patch projection layers for each input modality.
Specifically, patches of each modality are projected to tokens using a different linear projection for each modality.
In this work, we use a 2D convolution to extract $16 \times 16$ patches and project them to the input dimension $D$.
Next, position embeddings are added to the projected vectors so that the model is able to localize and distinguish each embedded patch.
In addition to the multimodal input data, the learnable fusion tokens are introduced as one of the inputs.
Different to the bottleneck fusion tokens in MBT and Zrror, we use the spatial tokens for dense downstream tasks, which have the same number of tokens of full input patches.
In order to get local features, we add 2D sine-cosine positional embeddings on the spatial fusion tokens and use Bi-LSTM to aggregate all modality information to fusion tokens.
Then the projected patches together with the learnable tokens are concatenated into a sequence of tokens and given as input to the same Transformer encoder with masked attention.
Since all our input data have a 2D structure, we add 2D sine-cosine positional embeddings after linear projection.
Following the setting of MultiMAE, we don't consider any modality-specific positional embedding.

\subsection{Bi-LSTM Attention}
We use a Bi-LSTM with an attention mechanism to integrate different modality input embeddings into learned fusion tokens for improving the feature learning ability.
Consider one direction of the LSTM network: let $\mathop{h_i}\limits ^{\rightarrow}$ be the output of the LSTM for the multimodal inputs (Optical, SAR, DEM, MAP) and the learned fusion tokens.
Bi-LSTM performs forward training and backward training separately for each training sequence and then combines the results of forward training and backward training together as the output of each modality, which is noted as $h_i = [\mathop{h_i}\limits ^{\rightarrow}, \mathop{h_i}\limits ^{\leftarrow}]$.
We use $h_{f}$ (fusion tokens) to attend to all multimodal inputs $h_{o}$ (optical tokens), $h_{s}$ (SAR tokens), $h_{d}$ (DEM tokens), $h_{m}$ (map tokens) and measure the importance of each modality through the similarity  with a learning parameter $u$.
Then we get a normalized importance weight $\beta_i$ through a softmax function.
\begin{equation}
	\beta_i=\frac{exp(\boldsymbol{u}^{\boldsymbol{\top}} \tanh \left(\boldsymbol{W}\left[\boldsymbol{h}_f ; \boldsymbol{h}_i\right] + b \right))}{\sum_{i=1}^{t-1} exp( \boldsymbol{u}^{\boldsymbol{\top}} \tanh \left(\boldsymbol{W}\left[\boldsymbol{h}_f ; \boldsymbol{h}_i\right] + b \right))}
\end{equation}
where $\boldsymbol{u}$ and $\boldsymbol{h}$ have the same dimension as the cell state of the LSTM, [] is the concatenate operation. $\boldsymbol{W}$ is a weight matrix of the MLP and $b$ is a bias vector of the MLP.
The final new fusion token is thus:
\begin{equation}
	\boldsymbol{a}=\sum_{i=1}^{t-1} \beta_i \cdot \boldsymbol{h}_i
\end{equation}

\subsection{Masked Self-Attention}
Masked Self-Attention is the key block of Multimodal Transformer in contrastive pre-training.
Using masked attention, we force part of the representation to attend only to itself, while other parts can attend to the whole representation.
The main goal of this approach is to split the representation into five parts: a part which only focuses on Optical tokens, a part which focuses on SAR tokens, a part which focuses on DEM tokens, a part which focuses on MAP tokens, and the fusion tokens which can attend to the whole representation.
In this architecture, the self-attention in each layer and the cross-attention in the last layer both used this masking strategy.
Here we introduce the masking binary tensor m that specifies which vectors can access each other.
Entries of the masking matrix are $m_{i,j} = 1$ if information can flow from latent $j$ to latent $i$.
Versus, we set $m_{i,j} = 0$.
The mask is applied to the standard attention output operation can be expressed as :
\begin{equation}
	o_i=\sum_j \hat{a}_{i j} \cdot v_j;~~where ~ \hat{a}_{i j}=\frac{m_{i j} \exp \left(\frac{q_i^{\top} k_j}{\sqrt{D}}\right)}{\sum_{\left\{j^{\prime}, m_{i j^{\prime}}=1\right\}} \exp \left(\frac{q_i^{\top} k_{j^{\prime}}}{\sqrt{D}}\right)}
\end{equation}
In order to keep the performance of a single modality when other modalities are absent, the modality-specific representation can not access the fusion representation or other modalities.
This explicitly prevents the information of the fusion stream from leaking into the unimodal representation.
This is the key to preserving pure streams that correspond to single modalities.
For example, after applying this mask, the SAR-specific output $o_{s}$ only contains information coming from the SAR input.
The optical-specific output $o_{o}$ only contains information coming from the optical input.
The DEM-specific output $o_{d}$ only contains information coming from the DEM input.
The MAP-specific output $o_{m}$ only contains information coming from the MAP input.
The fusion output $o_f$ access all outputs in the model.

\subsection{Reconstruction Pretraining}
In order to train our network in an MAE way, we use a separate decoder for each generation task.
The input to each decoder is the spatial tokens output from the cross attention.
Following the same setting of MAE, we use shallow decoders with a low dimensionality which consists of two Transformer blocks.
MultiMAE mask across different modalities ensures the model develops predictive coding across different modalities besides different spatial patches.
According to MultiMAE, we set a constant number of visible tokens at 256, which corresponds to 1/4 of all tokens in our experiment (learned fusion tokens and three modality inputs with 256 $\times$ 256 image size and 16 $\times$ 16 patch size).
The proportion of tokens per modality $\lambda$ by sampling from a symmetric Dirichlet distribution $(\lambda_{Optical}, \lambda_{SAR}, \lambda_{DEM}, \lambda_{MAP}) \sim Dir(\alpha)$, where $\lambda_{Optical} + \lambda_{SAR} + \lambda_{DEM} + \lambda_{MAP} = 1, \lambda \geq 0$.
For simplicity and better representation of any possible sampled task, we use a concentration parameter $\alpha = 1$.
As shown in Fig. \ref{fig6-1}, we adopt reconstruction loss ($l_1$ distance Mean Squared Error) to respectively recover the pixel color and height information following MultiMAE and using cross-entropy loss ($l_{ce}$) on land-cover map reconstruction:
\begin{equation}
	\begin{array}{l}
		L_{DEM}=l_1(Dec(o_f), DEM) \\
		L_{SAR\_RGB}=l_2(Dec(o_f), SAR) + l_2(Dec(o_f), RGB)\\
		L_{MAP} = l_{ce}(Dec(o_f), MAP)
	\end{array}
\end{equation}

\subsection{Contrastive Pretraining}
We also add the class token for each modality input data and an additional global class token for learned fusion tokens. 
To integrate information from the encoded visible tokens of other modalities, we add a single cross-attention layer using these tokens as queries that cross-attend to the encoded tokens of the last self-attention layer.
We utilise the standard cross-attention operation and produce four different outputs: the vector outputs for each modality and a fusion vector output.
This design opens the possibility for contrastive learning among different modalities and fusion tokens.
For better multimodality alignment, we propose to use extra contrastive loss between each modality-specific output and the fusion vector.
Specificaly, given the Optical vector output $z_{o} = g_{o}(o_{o})$ and the fusion output $z_{f} = g_{f}(o_{f})$, where $g_{o}$ and $g_{o}$ are the linear projection for each modality.
The contrastive loss can be formulated as:
\begin{equation}\label{eq3}
	L_c(z_{o}, z_{f})=-\underset{S}{\mathbb{E}}{\left[\log \frac {e^{sim(z_{o}^i, z_{f}^i)/\tau}}{{\sum_{j=1}^{N}} e ^{sim(z_{o}^i, z_{f}^j)/\tau}}\right]}
\end{equation}
where $sim$ is a similarity function (i.e., cosine similarity), $S$ is a set that contains $N-1$ negative samples and one positive sample.
This equation introduces the loss for RGB-FUSION contrastive training.
In order to contrast the output of all outputs, we define a contrastive loss between unimodal representations and fusion representations.
Finally, we can write the full loss as:
\begin{equation}
	\begin{split}
	L=&L_{DEM} + L_{SAR\_RGB} + \lambda_2 * (L_c(z_{f}, z_{o}) \\
	& + L_c(z_{f}, z_{s}) + L_c(z_{f}, z_{d}) + L_c(z_{f}, z_{m}))
	\end{split}
\end{equation}

\subsection{Random Modalities Combination}
Besides the network design, the training strategy is vital to the performance of modal-incomplete inputs.
The research \cite{ma2022multimodal} finds that the Transformer models tend to overfit to dominate modalities in tasks.
To improve the robustness of the proposed approach against modal-incomplete data, we propose to leverage a random modality combination training strategy.
Thanks to the proposed approach, we can randomly choose the different modality combinations or unimodal data in pretraining or supervised training on downstream tasks.
Because the proposed approach fuses all modalities using additional learned tokens, which greatly reduces the effects of modal-incomplete input.

\section{Experiments}
In this section, we evaluate the proposed approach in multiple settings.
We first introduce the multimodal dataset used in this work.
Then, we present the details of pre-training and the details of training on downstream tasks, as well as the evaluation procedures.
Finally, we ablate the performance of the complete and the incomplete multimodal input that show the proposed approach's flexibility.

\subsection{Experimental Details}
In order to showcase the proposed approach across the different modalities, we train the proposed approach in a completely supervised paradigm and a fine-tuning paradigm with a pre-trained weight.
Many works have pointed out that the pretraining of the big model on multimodal can be beneficial on downstream tasks \cite{singh2023effectiveness}.
And, the pre-trained model can be used for arbitrary downstream tasks with the finetuning of the task-specific decoder.
Hence we can train a giant model on a large multimodal data set with as many modalities as possible.
The pre-trained model can strengthen the ability to extract features that are only trained on a few or single modality data.
In this section, we provide the details of the self-supervised pre-training and the supervised training on downstream tasks as well as the multimodal datasets.

\begin{figure*}[pt]
	\centering
	\includegraphics[width=7.0 in]{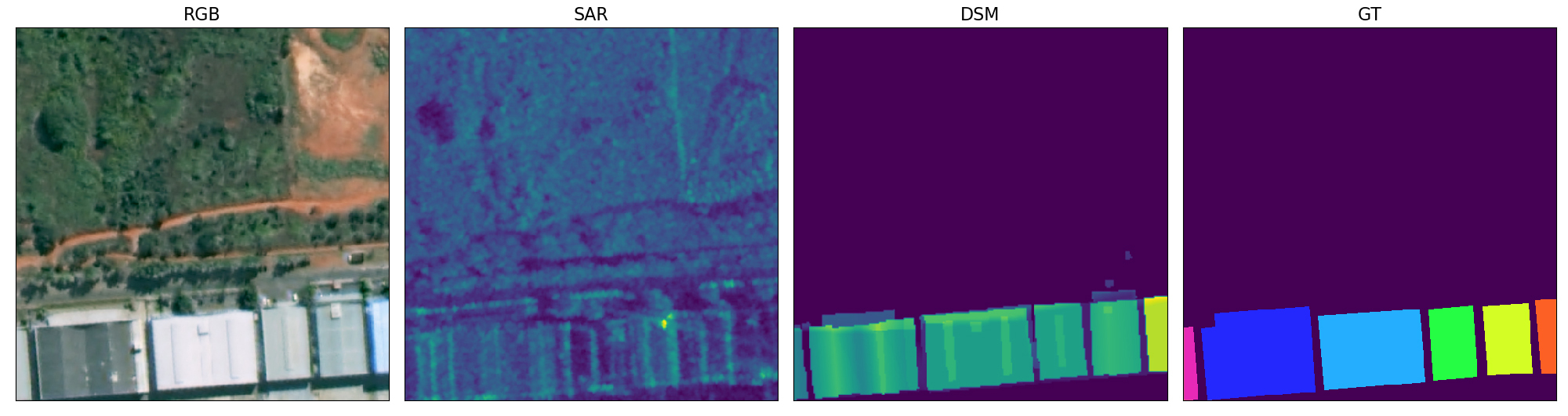}
	\caption{DFC2023 track2 data sample.}
	\label{fig6-2}
\end{figure*}

\subsection{Description of Datasets}
We train and test the performance of the proposed approach on two multimodal datasets for two downstream tasks, namely building instance / semantic segmentation and LULC mapping.
\subsubsection{DFC2023 track2: Building instance / semantic segmentation}
The former is performed on the track 2 dataset of DFC2023, which comprises a combination of RGB images, Synthetic Aperture Radar (SAR) images, and Digital Surface Model (DSM) data.
While the objective of the original task is building height estimation, this study simplifies it as building instance / semantic segmentation.
The dataset consists of images obtained from GaoJing-1, GaoFen-2 and GaoFen-3 satellites, with respective spatial resolutions of 0.5 m, 0.8 m and 1 m.
Normalized Digital Surface Models (nDSMs) are used as a reference in Track2 and are created from stereo images captured by GaoFen-7 and WorldView-1 and -2 with approximately 2 m ground sampling distance (GSD).
The dataset was collected from seventeen cities across six continents and hence is highly diverse in terms of landforms, building types and architecture.
The labels of building instance segmentation adopt the MS COCO format and are provided in a JSON file.
A sample of the labels is shown in Fig. \ref{fig6-2} for illustration.

\subsubsection{Quadruplets Dataset: Land-Use Land-Cover (LULC) mapping}
\begin{figure*}[pt]
	\centering
	\includegraphics[width=7.0 in]{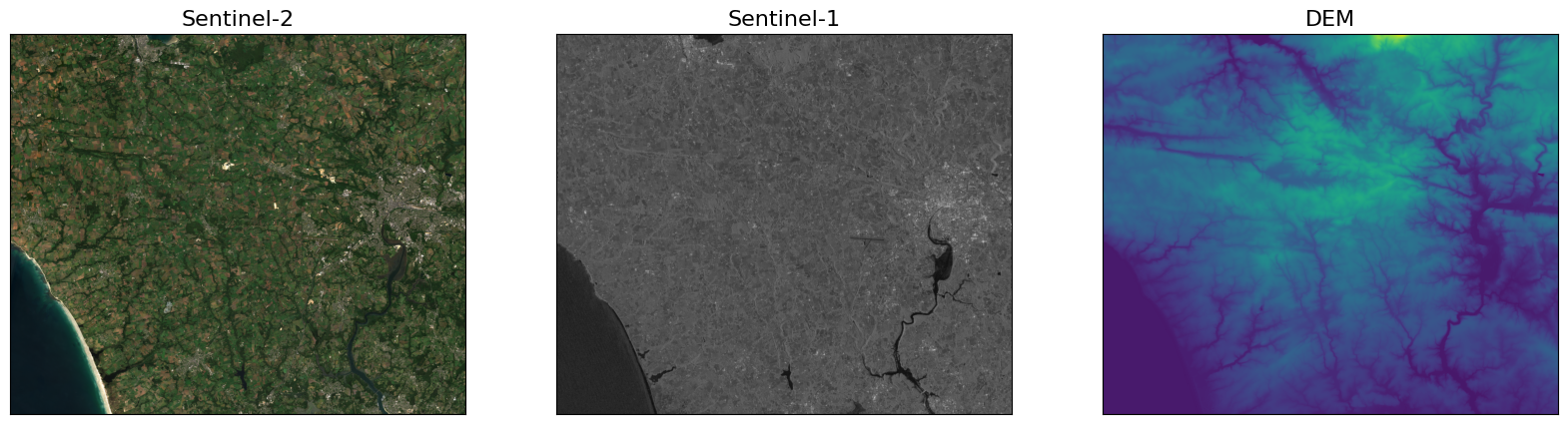}
	\caption{Sentinel1, Sentinel-2 and DEM data.}
	\label{fig6-3}
\end{figure*}

\begin{figure*}[pt]
	\centering
	\includegraphics[width=7.0 in]{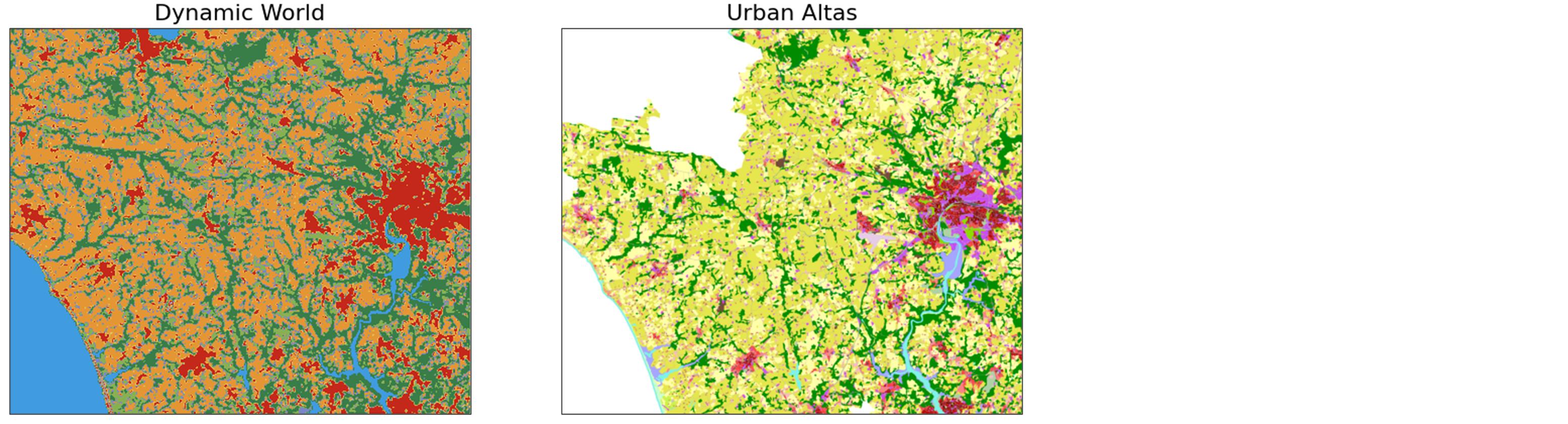}
	\caption{Dynamic World Map and European Urban Atlas data.}
	\label{fig6-4}
\end{figure*}

In the second dataset, we utilize diverse data sources obtained from Google Earth Engine (GEE) platform, encompassing Sentinel-1, Sentinel-2, Lidar DEMs and Dynamic World LULC maps, as shown in Fig. \ref{fig6-3} and Fig. \ref{fig6-4}.
The dataset comprises 37 regions across various landscapes and LULC classes in France and Australia.
The Sentinel-1 mission provides data from a dual-polarization C-band SAR instrument and produces the calibrated and ortho-corrected S1 GRD products.
We downloaded the data from the COPERNICUS/S1\_GRD category on GEE, resampling it into 10 m resolution and using dual-band VV+VH. 
Similarly, we downloaded the Sentinel-2 data from the COPERNICUS/S2\_SR\_HARMONIZED category, which provides multispectral imaging with 13 spectral bands suitable for large-scale LULC mapping. 
We resampled the Sentinel-2 data into 10 m resolution, and use the RGBN bands in this work.
The two types of Lidar DEM are provided in this research.
In France, we utilized the RGE ALTI dataset, which is a digital elevation model (DEM) created using airborne lidar, with a pixel size of 1 m.
We resampled this dataset to 10 meters, and its vertical accuracy ranges from 0.2 m to 0.5 m with an average accuracy of 7 m in steep slope areas.
In Australia, we use the digital elevation model 5 m grid derived from 236 individual LiDAR surveys conducted between 2001 and 2015.
We compiled and resampled the available 5 m resolution LiDAR-derived DEMs using a neighbourhood-mean method to create 10 m resolution datasets for each survey area, which we used in this work.
The Dynamic World MAP (DNW) dataset comprises globally consistent, 10 m resolution, near real-time land-use and land cover predictions derived from Sentinel-2 imagery. It features ten bands that include estimated probabilities for each of the nine LULC classes (water, trees, grass, crops, shrub and scrub, flooded vegetation, built-up area, bare ground, and snow \& ice). It also has a class "label" band indicating the class with the highest estimated probability, which makes it suitable for multi-temporal analysis and custom product creation.
Lastly, we utilized the labelled class-reference from the UrbanAtlas 2018 database containing 27 LULC classes as the label of this dataset. The dataset provides integer rasters with index labels.
We created raster maps with 10 m resolution that geographically match the Sentine-1/-2 images using the open-data vector images freely available on the European Copernicus program website.

\subsubsection{Downstream Tasks}
We evaluate the proposed approach against state-of-the-art methods on two downstream tasks: building instance / semantic segmentation, and the LULC mapping.
In particular, the evaluation is performed on the supervised learning and finetuning two paradigms.
For these two downstream tasks, we replace the pre-trained decoders with randomly initialized Mask2Former.
In the following, we give an overview of two tasks.

\textbf{Building Instance / Semantic Segmentation:} We follow the Mask2Former but replace the backbone with the proposed network.
In a supervised fashion, we train the whole network from scratch using a random modality combination strategy.
In the finetuning fashion, we provide two ways, one is only to update the network on the pre-trained ViT-B backbones using a generative way, and the other is to update the whole network on the pre-trained ViT-B backbones using reconstruction and contrastive losses.
We train our model on DFC2023 track2 train split and report the validation accuracy on the validation split.
Along with the result of building instance segmentation, we also provide the binary building semantic segmentation results.

\textbf{Land Use Land Cover Mapping:} 
We still use the Mask2Former with the proposed backbone on the quadruplets dataset to perform LULC maps. 
However, we considered seven classes merged from the semantic hierarchy defined by UrbanAtlas.
For that, we extract 7 semantic classes by taking the argmax of the prediction head.
Same training strategy as the building instance segmentation is used in this task.
We train our model on 10 (5340 samples) cities and report the validation accuracy on the other 2 (783 samples) cities.

\subsubsection{Architectural Details}
The proposed approach uses a ViT-B as the main structure consists of 4 and 5 input adapters with a patch size of 16$\times$16 pixels for the pre-training in two different tasks.
Different from standard MultiMAE, we add the learnable fusion tokens as input which use an additional input adapter to add 2D Sine-Cosin position Encoding.
The fusion tokens have as many as the number of patched inputs of each modality.

After adding the position encodings, the fusion tokens with all modality inputs are input into a one-layer Bi-LSTM attention block.
In self-attention, we use the masked algorithm to avoid the fusion information leak to a single modality.
In order to get the global feature of each modality and the fusion output, we use an additional cross-attention layer to map the patch embeddings into the vector output.
Then an auxiliary contrastive loss is added between each modality output vector and fusion output vector.

For reconstruction learning, we follow the same setting of the MultiMAE decoder but without positional embeddings and cross-attention layer.
The fusion tokens are projected to the decoder dimension by using a linear projection layer and then added a learned modality embedding.
After these, two Transformer blocks and a linear projector are used to project and reshape it to form an image or map.

For two downstream tasks, we adopt the same settings from Mask2Former.
For the pixel decoder, we use 6 MSDeformAttn layers applied to feature maps with resolution 1/8, 1/16 and 1/32, and use a simple upsampling layer with lateral connection on the final 1/8 feature map to generate the feature map of resolution 1/4 as the per-pixel embedding.
We use the Transformer decoder with 9 layers and 100 queries for instance segmentation, 9 queries for binary building semantic segmentation and 9 queries for LULC mapping.
We use the binary cross-entropy loss and the dice loss for the mask loss. The final loss is a combination of mask loss and classification loss.
For instance segmentation, we use the standard AP@50 metric. For semantic segmentation, we use mIoU (mean Intersection-over-Union).

\subsubsection{Training Details}
For pre-training, we train our model for 1600 epochs on 5700 triplet data on DFC2023 and 6123 quadruplet data, individually.
We use the AdamW optimizer with a base learning rate of 1e-4 and weight decay of 0.05.
We warm up training for 40 epochs, starting from using cosine decay.
We set the batch to 40 using a single 3090.
All data are resized to 256$\times$256.
The number of non-masked tokens given to the encoder is set to 256 on two data sets.
For the second dataset, where we use the land-cover map as an additional modality input with 64-dimensional class embeddings.

For instance segmentation and semantic segmentation using Mask2Former, we use AdamW optimizer and the step learning rate schedule.
We use an initial learning rate of 0.0001 and a weight decay of 0.05.
A learning rate multiplier of 0.1 is applied to the backbone with the pretraining and not in the supervised paradigm.
And, we decay the learning rate at 0.9 and 0.95 fractions of the total number of training steps by a factor of 10.
We train our models for 50 epochs with a batch size of 10 in the semantic segmentation task and 300 epochs in the instance segmentation task.

\subsection{Experimental Results}
\begin{table*}[pb]
	\centering
	\caption{Quantitative evaluations of proposed approach versus MultiViT with complete and incomplete multimodality inputs on the DFC2023 track2 dataset. Results are reported on AP@50 for instance segmentation and mIoU for semantic segmentation.}
	\label{tab6-1}
	\renewcommand\tabcolsep{14.0pt}
	\centering
	\begin{tabular}{ccccc|cccc}
		\hline
		\multirow{2}{*}{Multimodal Input} & \multicolumn{2}{c}{MultiViT} & \multicolumn{2}{c|}{Sup. Propsed} & \multicolumn{2}{c}{Fine. w/ G.} & \multicolumn{2}{c}{Fine. w/ G. \& C.} \\
		& ins. & sem.  & ins. & sem. & ins. & sem. & ins. & sem. \\ \hline
		SAR, RGB, DSM  & 0.147 & 0.820 & 0.333 & 0.851 & 0.298 & 0.852 & 0.300 & 0.849 \\ \hline
		SAR, RGB       & 0.002 & 0.523 & 0.296 & 0.809 & 0.257 & 0.797 & 0.260 & 0.798 \\
		SAR, DSM       & 0.064 & 0.700 & 0.233 & 0.779 & 0.217 & 0.776 & 0.202 & 0.780 \\
		RGB, DSM       & 0.105 & 0.736 & 0.332 & 0.847 & 0.298 & 0.848 & 0.300 & 0.844 \\ \hline
		SAR            & 0.001 & 0.392 & 0.040 & 0.552 & 0.036 & 0.532 & 0.037 & 0.566 \\
		RGB            & 0.003 & 0.457 & 0.291 & 0.799 & 0.252 & 0.788 & 0.254 & 0.784 \\
		DSM            & 0.036 & 0.683 & 0.211 & 0.753 & 0.200 & 0.754 & 0.187 & 0.754 \\ \hline
	\end{tabular}
\end{table*}

\subsubsection{Multimodal Comparison}
We evaluate the proposed approach by two paradigms, one is supervised from scratch, and the other is finetuning with pre-trained weights.
To evaluate the former, we compare the proposed approach against a technique that uses origin self-attention and does not employ the random modality combination training strategy, termed MultiViT, on modal-complete and modal-incomplete inputs for building instance/semantic segmentation and LULC mapping tasks. 
The results reported in Tables \ref{tab6-1} and \ref{tab6-2} reveal that the proposed approach outperforms MultiViT in building instance/semantic segmentation tasks when evaluated with modal-complete inputs.
However, for the LULC mapping task, the performance of the proposed approach and MultiViT are comparable. 
With regards to model-incomplete inputs, the proposed approach performs impressively on all modality incomplete inputs and single modality inputs for both tasks due to the proposed attention block and random modality combination training strategy. 
For building instance/semantic segmentation, there is a visible dominance of RGB images over all other modalities, followed by DSM, while SAR images make the slightest contribution to the task, even causing the noise. 
In this situation, MultiViT completely overfits on dominant modality inputs and fails on the task with single modality inputs when evaluated with model-incomplete inputs. 
Similarly, for LULC mapping, sentinel-2 images along with a dynamic world map have a significant influence on the task, followed by sentinel-1 and DEM images. 
The proposed approach achieves the best performance with the mIoU of 0.244 with modal-complete inputs while MultiViT overfits on dynamic world maps, and performs slightly better when a dynamic world map is present but fails altogether when it is not present in the inputs.

In the context of the finetuning paradigm, the proposed approach is assessed through two distinct pretraining methods: one that employs generative pretraining and another that combines generative and contrastive pretraining. 
The outcomes of the evaluation for both tasks are presented in Table \ref{tab6-1} and Table \ref{tab6-2}. 
As one can see, two tasks show controversial results.
Specifically, in the case of building instance/semantic segmentation tasks, the training-from-scratch model outperforms all other models. 
However, the model that leverages both generative and contrastive pretraining methods is closely ranked as the second-best. 
In contrast, for the land-cover mapping task, the fully finetuned model is the top-performing model among all the models listed in the tables, demonstrating the potential of pre-training in augmenting downstream LULC tasks.



\begin{figure*}[pt]
	\centering
	\includegraphics[width=6.5 in]{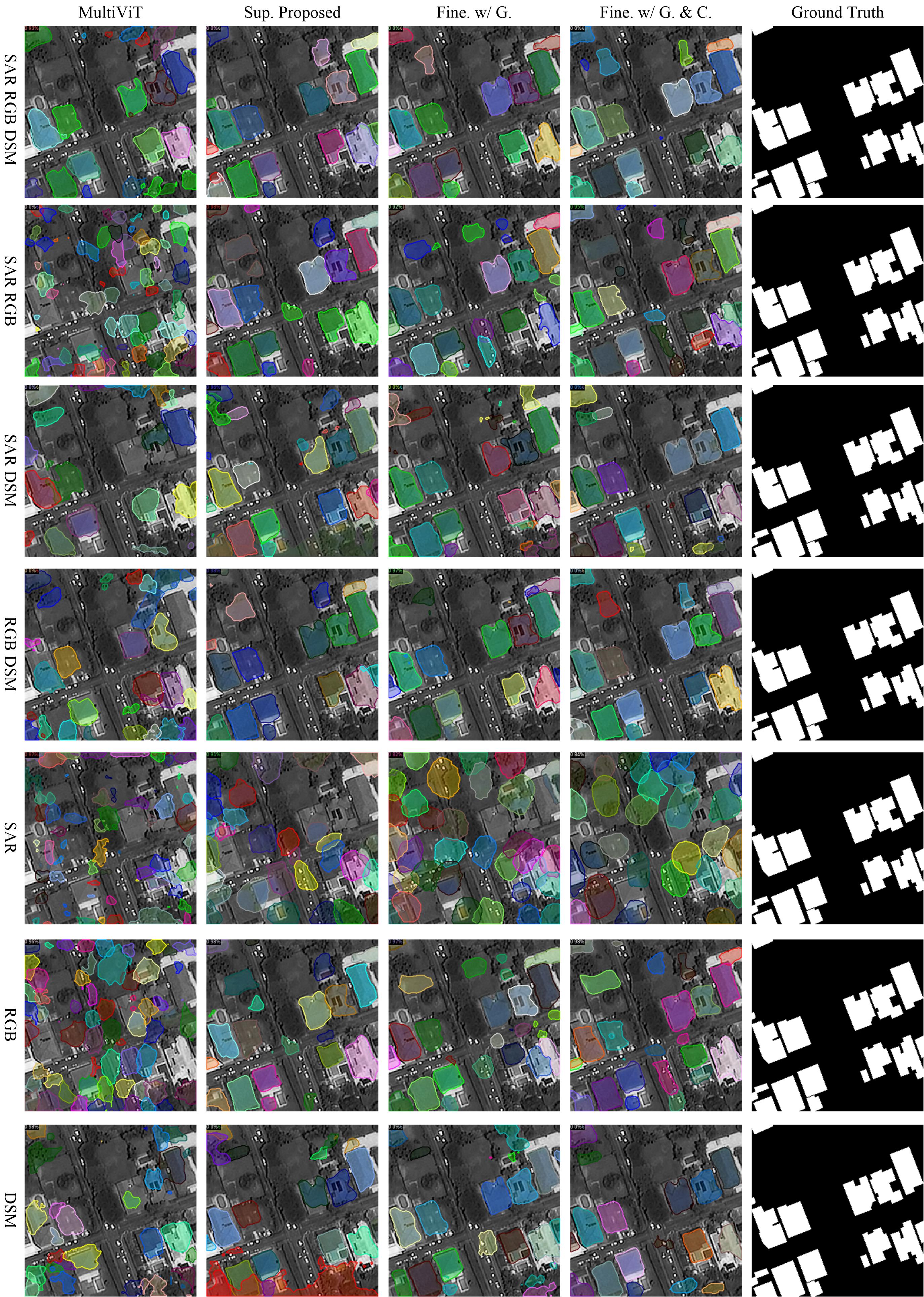}
	\caption{Results of proposed approaches in supervised paradigm and two finetuning paradigms versus MultiViT on DFC2023 track2 dataset.}
	\label{fig6-5}
\end{figure*}

\begin{figure*}[pt]
	\centering
	\includegraphics[width=6.5 in]{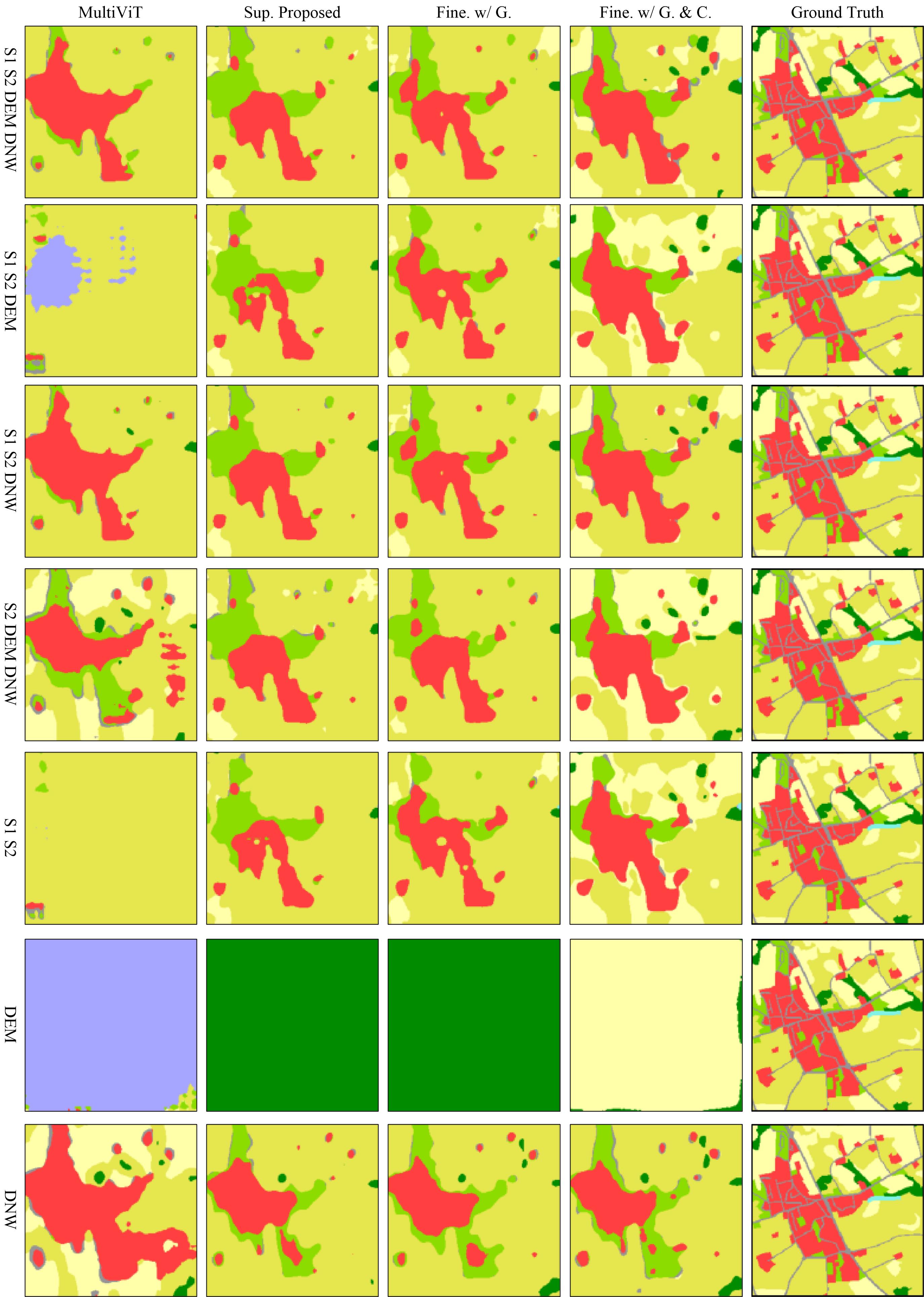}
	\caption{Results of proposed approaches in supervised paradigm and two finetuning paradigms versus MultiViT on the quadruplets dataset.}
	\label{fig6-6}
\end{figure*}

For the single modality input, our goal is not to show state-of-the-art performance in this setting, as we are trying to solve the dramatic degradation of unimodal inference with a multimodal backbone.
Here we show the ability of the proposed approach to producing meaningful unimodal outputs when fed with unimodal data.
To do this, we only input one modality and miss out on other modality inputs.
As we can see from both datasets (Table \ref{tab6-1} and Table \ref{tab6-2}), the MultiViT suffers significant degradation from the modal missing and completely fails to work on the non-dominated modalities.
In contrast, the proposed approach using the random modality combination strategy achieves high performance when only one modality is available.
This is due to the fact that in those models, some capacity is allocated to each modality specifically and the model is able to produce unimodal outputs.
Besides the quantitative analysis, we also provide a visual qualitative comparison, where Fig. \ref{fig6-2} and Fig. \ref{fig6-3} show the results of building instance / semantic segmentation and LULC mapping.
For building instance / semantic segmentation, similar to Table \ref{tab6-1}, the proposed approach with supervised paradigm achieved the best performance, then is the results of finetuning.
The MultiViT achieves the worst performance, especially with the modal-incomplete inputs.
Our experimental results reveal that the SAR modality produced inferior results compared to other modalities.
For the LULC mapping task, the finetuning with contrastive and generative pre-trained weights outperformed other approaches, while MultiViT exhibited reliable performance only with DNW input.
For different modalities, we conclude that the Sentinel-1/2 images and DNW maps contributed equally as effective modalities, while the DEM input was determined to be a single-class predictor, indicating its inability to extract useful information.

\begin{table*}[pt]
	\centering
	\caption{Quantitative evaluations of proposed approach versus MultiViT with complete and incomplete multimodality inputs on the quadruplets dataset. The results are reported on mIoU.}
	\label{tab6-2}
	\renewcommand\tabcolsep{14.0pt}
	\centering
	\begin{tabular}{ccc|cc}
		\hline
		Multimodal Input & MultiViT & Sup. Proposed & Fine. w/ G. & Fine. w/ G. \& C. \\ \hline 
		S1, S2, DEM, DNW & 0.222 & 0.244 & 0.243 & 0.246  \\ \hline
		S1, S2, DEM      & 0.070 & 0.229 & 0.235 & 0.238  \\
		S1, S2, DNW      & 0.219 & 0.244 & 0.243 & 0.246  \\
		S1, DEM, DNW     & 0.219 & 0.235 & 0.235 & 0.235  \\
		S2, DEM, DNW     & 0.223 & 0.237 & 0.230 & 0.240  \\ \hline
		S1, S2           & 0.069 & 0.232 & 0.235 & 0.240  \\
		S1, DEM          & 0.074 & 0.208 & 0.221 & 0.216  \\
		S1, DNW          & 0.219 & 0.239 & 0.236 & 0.235  \\
		S2, DEM          & 0.054 & 0.210 & 0.204 & 0.227  \\
		S2, DNW          & 0.217 & 0.239 & 0.232 & 0.239  \\
		DEM, DNW         & 0.209 & 0.234 & 0.227 & 0.238  \\ \hline
		S1               & 0.079 & 0.208 & 0.222 & 0.214  \\
		S2               & 0.062 & 0.215 & 0.210 & 0.228  \\
		DEM              & 0.015 & 0.010 & 0.013 & 0.051  \\
		DNW              & 0.207 & 0.234 & 0.226 & 0.237  \\ \hline
	\end{tabular}
\end{table*}

\subsubsection{Albation}
We now analyze the proposed approach through a series of ablation studies on both finetuning and supervised paradigms.
To evaluate the generalizability of the proposed components, all ablations were performed on both tasks: the building instance / semantic segmentation and LULC mapping.

\textbf{Random Modality Combination \& Bi-LSTM Attention.}
We first validate the importance of the modality random combination training strategy on downstream tasks in a supervised paradigm.
As shown in Table \ref{tab6-3} and Table \ref{tab6-4}, the model without the modality random combination training strategy experienced severe degradation with modal-incomplete inputs and even failed with a single modality on both tasks. 
In addition, we test the effect of the Bi-LSTM attention by removing it from the proposed network.
The corresponding results showed a significant drop in performance, indicating that the Bi-LSTM enables superior interaction of the fusion token with each modality and facilitates learning more discriminative features for downstream tasks.

\textbf{Partial Fine-tuning and Non-masked Attention.} 
In addition to the finetuning of whole model, partial finetuning is also used to evaluate the quality of the learned representation in a self-supervised approach.
Partial finetuning involves freezing the backbone and updating only the task-specific decoder on two tasks. 
It is important to note that contrastive pre-training relies on masked attention to keep each modality independent, especially when working with different data formats such as text and images. 
The use of masked attention in contrastive pre-training helps in avoiding information flow from one modality to the other, thereby keeping modality-specific information through the network. This is more beneficial for downstream tasks that involve only a single modality. 
However, when using generative pre-training, masked self-attention is not mandatory. 
Here, we show the finetuning result based on the combination pretraining (the use of reconstruction loss and contrastive loss), the generative pretraining (the only use of reconstruction loss), and the finetuning result without masked self-attention.
The results are shown in Table \ref{tab6-5} and Table \ref{tab6-6} for both tasks.
In the first row, we remove the masked Self-Attention blocks while keeping the random modality combination training strategy in finetuning, which results in significant improvement in performance.
This is probably because masked self-attention hinders the interaction between different modalities. 
Compared with the generative pre-training, the use of masked attention in the combination pre-training helps to avoid the information flow from one modality to the other.
As one can see, the unimodal inference performs close to the modal-incomplete inputs as the modality streams are more independently treated.
In contrast, the results without contrastive pre-training tend to overfit on dominant modalities and perform relatively poorly on other modalities.
And, it further faces lower performance on one single modality.

\begin{table}[pt]
	\centering
	\caption{Quantitative evaluations of the proposed approach on the different settings of Bi-LSTM and random modality combination training strategy with complete and incomplete multimodality inputs on the DFC2023 track2 dataset. Results are reported on AP@50 for instance segmentation and mIoU for semantic segmentation.}
	\label{tab6-3}
	\renewcommand\tabcolsep{4pt}
	\centering
	\begin{tabular}{ccccccc}
		\hline
		\multirow{2}{*}{Multimodal Input} & \multicolumn{2}{c}{Sup. w/o LSTM} & \multicolumn{2}{c}{Sup. w/o Random} & \multicolumn{2}{c}{Sup. w/ all} \\
		& ins. & seg. & ins. & seg. & ins. & seg. \\ \hline
		SAR, RGB, DSM  & 0.265 & 0.809 & 0.301 & 0.854 & 0.333 & 0.851 \\ \hline
		SAR, RGB       & 0.213 & 0.728 & 0.083 & 0.660 & 0.296 & 0.809 \\
		SAR, DSM       & 0.173 & 0.763 & 0.061 & 0.696 & 0.233 & 0.779 \\
		RGB, DSM       & 0.165 & 0.807 & 0.224 & 0.782 & 0.332 & 0.847 \\ \hline
		SAR            & 0.028 & 0.509 & 0.000 & 0.372 & 0.040 & 0.552 \\
		RGB            & 0.210 & 0.722 & 0.061 & 0.577 & 0.291 & 0.799 \\
		DSM            & 0.168 & 0.749 & 0.040 & 0.664 & 0.211 & 0.753 \\ \hline
	\end{tabular}
\end{table}

\begin{table}[pt]
	\centering
	\caption{Quantitative evaluations of the proposed approach on the different settings of Bi-LSTM and random modality combination training strategy with complete and incomplete multimodality inputs on the quadruplets dataset. The results are reported on mIoU.}
	\label{tab6-4}
	\renewcommand\tabcolsep{4.0pt}
	\centering
	\begin{tabular}{cccc}
		\hline
		Multimodal Input & Sup. w/o LSTM & Sup. w/o Random & Sup. w/ all \\ \hline
		S1, S2, DEM, DNW & 0.242 & 0.244 & 0.244 \\ \hline
		S1, S2, DEM      & 0.227 & 0.175 & 0.229 \\
		S1, S2, DNW      & 0.244 & 0.247 & 0.244 \\
		S1, DEM, DNW     & 0.237 & 0.198 & 0.235 \\
		S2, DEM, DNW     & 0.240 & 0.239 & 0.237 \\ \hline
		S1, S2           & 0.228 & 0.174 & 0.232 \\
		S1, DEM          & 0.201 & 0.058 & 0.208 \\
		S1, DNW          & 0.239 & 0.197 & 0.239 \\
		S2, DEM          & 0.211 & 0.139 & 0.210 \\
		S2, DNW          & 0.241 & 0.239 & 0.239 \\
		DEM, DNW         & 0.231 & 0.179 & 0.234 \\ \hline
		S1               & 0.203 & 0.051 & 0.208 \\
		S2               & 0.212 & 0.136 & 0.215 \\
		DEM              & 0.013 & 0.053 & 0.010 \\
		DNW              & 0.233 & 0.163 & 0.234 \\ \hline
	\end{tabular}
\end{table}

\begin{table}[pt]
	\centering
	\caption{Quantitative evaluations of the proposed approach in finetuning paradigm with different settings with complete and incomplete multimodality inputs on DFC2023 track2 dataset. Results are reported on AP@50 for instance segmentation and mIoU for semantic segmentation.}
	\label{tab6-5}
	\renewcommand\tabcolsep{4pt}
	\centering
	\begin{tabular}{ccccccc}
		\hline
		\multirow{2}{*}{Multimodal Input} & \multicolumn{2}{c}{Fine. w/o Mask} & \multicolumn{2}{c}{Partial Fine.} & \multicolumn{2}{c}{Full Fine.} \\
		& ins. & seg. & ins. & seg. & ins. & seg. \\ \hline
		SAR, RGB, DSM & 0.317 & 0.850 & 0.215 & 0.807 & 0.300 & 0.849 \\ \hline
		SAR, RGB      & 0.276 & 0.799 & 0.136 & 0.711 & 0.260 & 0.798 \\
		SAR, DSM      & 0.220 & 0.783 & 0.173 & 0.767 & 0.202 & 0.780 \\
		RGB, DSM      & 0.318 & 0.845 & 0.206 & 0.800 & 0.300 & 0.844 \\ \hline
		SAR           & 0.034 & 0.562 & 0.022 & 0.499 & 0.037 & 0.566 \\
		RGB           & 0.276 & 0.789 & 0.132 & 0.694 & 0.254 & 0.784 \\
		DSM           & 0.205 & 0.752 & 0.152 & 0.747 & 0.187 & 0.754 \\ \hline
	\end{tabular}
\end{table}

\begin{table}[pt]
	\centering
	\caption{Quantitative evaluations of the proposed approach in finetuning paradigm with different settings with complete and incomplete multimodality inputs on the quadruplets dataset. The results are reported on mIoU.}
	\label{tab6-6}
	\renewcommand\tabcolsep{4pt}
	\centering
	\begin{tabular}{cccc}
		\hline
		Multimodal Input & Fine. w/o Mask & Partial Fine. & Full Fine. \\ \hline
		S1, S2, DEM, DNW & 0.250 & 0.233 & 0.246  \\ \hline
		S1, S2, DEM      & 0.243 & 0.223 & 0.238  \\
		S1, S2, DNW      & 0.248 & 0.222 & 0.246  \\
		S1, DEM, DNW     & 0.238 & 0.217 & 0.235  \\
		S2, DEM, DNW     & 0.242 & 0.221 & 0.240  \\ \hline
		S1, S2           & 0.238 & 0.223 & 0.240  \\
		S1, DEM          & 0.221 & 0.212 & 0.216  \\
		S1, DNW          & 0.240 & 0.224 & 0.235  \\
		S2, DEM          & 0.231 & 0.198 & 0.227  \\
		S2, DNW          & 0.242 & 0.221 & 0.239  \\
		DEM, DNW         & 0.245 & 0.216 & 0.238  \\ \hline
		S1               & 0.226 & 0.212 & 0.214  \\
		S2               & 0.239 & 0.203 & 0.228  \\
		DEM              & 0.023 & 0.011 & 0.051  \\
		DNW              & 0.241 & 0.214 & 0.237  \\ \hline
	\end{tabular}
\end{table}

\section{Conclusion}
In this work, we introduce an incomplete multimodal learning framework for multimodal remote sensing data fusion which can be used in both supervised training and self-supervised pretraining paradigms.
Unlike previous multimodal remote sensing data fusion approaches, the proposed approach enables the training and inference of models with modal-incomplete inputs. 
By using the Bi-LSTM attention mechanism and masked self-attention, we are able to pre-train the network using contrastive and reconstruction losses in the MultiMAE framework, and also to train the network from scratch or finetune the model on downstream tasks using a random modality combination strategy.
This strategy allows the network to maintain performance even when dealing with modal-incomplete inputs or a single modality in the inference stage. 
We evaluate our model on two multimodal remote sensing datasets, demonstrating flexibility in network training and inference, and state-of-the-art performance when presented with modal-incomplete inputs.
However, this study focuses solely on different modality raster data. In future work, diverse modalities such as text and vector data will be incorporated into the framework.


\ifCLASSOPTIONcaptionsoff
  \newpage
\fi

\bibliographystyle{IEEEtran}
\bibliography{bibtex/mylib}

\end{document}